\documentclass[11pt,a4paper]{article}
\pdfoutput=1
\usepackage[nohyperref]{acl2017}
\usepackage{times}
\usepackage{latexsym}

\usepackage{url}
\usepackage{arydshln} 
\usepackage{color}

\usepackage[pdftex]{graphicx}

\aclfinalcopy 
\def\aclpaperid{93} 

\title{Predicting Causes of Reformulation in Intelligent Assistants}

\author{Shumpei Sano, Nobuhiro Kaji, and Manabu Sassano\\
	  Yahoo Japan Corporation\\
	  1-3 Kioicho, Chiyoda-ku, Tokyo 102-8282, Japan\\
        {\tt \{shsano, nkaji, msassano\}@yahoo-corp.jp}}

\date{}

\begin{document}
\maketitle
\begin{abstract}
  Intelligent assistants (IAs) such as Siri and Cortana conversationally interact with users and execute a wide range of actions (e.g., searching the Web, setting alarms, and chatting). IAs can support these actions through the combination of various components such as automatic speech recognition, natural language understanding, and language generation. However, the complexity of these components hinders developers from determining which component causes an error. To remove this hindrance, we focus on reformulation, which is a useful signal of user dissatisfaction, and propose a method to predict the reformulation causes. We evaluate the method using the user logs of a commercial IA. The experimental results have demonstrated that features designed to detect the error of a specific component improve the performance of reformulation cause detection.
\end{abstract}

\section{Introduction}
Intelligent assistants (IAs) such as Apple's Siri and Cortana have gained considerable attention as mobile devices have become prevalent in our daily lives. They are hybrids of search and dialogue systems that conversationally interact with users and execute a wide range of actions (e.g., searching the Web, setting alarms, making phone calls and chatting). IAs can support these actions through the combination of various components such as automatic speech recognition (ASR), natural language understanding (NLU), and language generation (LG).
One major concern in the development of commercial IAs is how to speed up the cycle of the system performance enhancement. The enhancement process is often performed by manually investigating user logs, finding erroneous data, detecting the component responsible for that error, and updating the component. As IAs are composed of various components, the error cause detection becomes an obstacle in the manual process. In this paper, we attempt to automate the error cause detection to overcome this obstacle.

One approach to do this is to utilize user feedback. In this work, we focus on reformulation, i.e., when a user modifies the previous input. In web search and dialogue systems, reformulation is known as an implicit feedback signal that the user could not receive a desired response to the previous input due to one or more system components failing.
In IAs, ASR error is a major cause of reformulation and has been extensively studied \cite{Hassan15,Schmitt15}.

Besides correcting ASR errors, users of IAs reformulate their previous utterances when they encounter NLU errors, LG errors, and so on. For example, when a user utters ``alarm'', the NLU component may mistakenly conclude that s/he wants to perform a Web search, and consequently the system shows the search results for ``alarm.'' \newcite{Sarikaya17} reported that only 12\% of the errors in an IA system are related to ASR components, which is the smallest percentage across six components. They also reported that the NLU component is the biggest source of errors (24\%). Therefore, the errors related to the other components should not be ignored to improve system performance. However, previous work mainly focused on reformulation caused by ASR error, and reformulation caused by the other components has received little attention.

In this work, we propose a method to predict reformulation causes, i.e., detect the component responsible for causing the reformulation in IAs. Features are divided into several categories mainly on the basis of their relations with the components in an IA. The experiments demonstrate that these features can improve the error detection performances of corresponding components. The proposed method which combines all features sets, outperforms the baseline, which uses component-independent features such as session information and reformulation related information.

Our work makes the following contributions. First, we investigate the reformulation causes among the components in IAs from real data of a commercial IA. Second, we create dataset of human annotated data obtained from a commercial IA. Finally, we develop the method to predict reformulation causes in IAs.

\section{Related Work}
Three research areas are related to our work, and the most closely related is reformulation (also called correction). As reformulation is frequently caused by system errors, the second related area is error analysis and error detection in search or dialogue systems. The third area is system evaluation in search or dialogue systems. Reformulation is a useful feature for system evaluation.

\subsection{Reformulation and Correction}
Users of search or dialogue systems often reformulate their previous inputs when trying to obtain better results \cite{Hassan13} or correct errors (e.g., ASR errors) \cite{Jiang13, Hassan15}. Our research focuses on the latter category of reformulation, which relates to correction. Studies on correction have mainly focused on automatic detection of correction \cite{Levow98,Hirschberg01a,Litman06}.
Some studies have also tried to improve system performance beyond correction detection. \newcite{Shokouhi16} constructed a large-scale dataset of correction pairs from search logs and showed that the database enables ASR performance to be improved.

The research most related to ours is that of \newcite{Hassan15}.
In addition to reformulation detection, they proposed a method to detect whether a reformulation is caused by ASR error or not. We extend their study to determine which component (e.g., ASR, NLU, and LG) is responsible for the error.
%
%
\subsection{Error Analysis and Error Detection}
Besides reformulation, researchers have studied system errors in search or dialogue systems. For example, \newcite{Meena15} and \newcite{Hirst94} focused on miscommunication in spoken dialogue systems and \newcite{Feild10} focused on user frustrations in search systems. In this paper, we focus on predicting the cause of errors among the different components in IAs. In spoken dialogue systems, \newcite{Georgiladakis16} reported that ASR error is the most frequent cause of errors among seven components (65.9\% in Let's Go datasets \cite{Raux05}. On the other hand, \newcite{Sarikaya17} reported that ASR error is the least frequent cause of errors among six components (12\% in an IA). These results indicate that errors in IA have various causes and that the causes other than ASR error also should not be ignored. In this paper, we focus on reformulation and propose a method to automatically detect the reformulation causes in IAs. 
\subsection{System Evaluation}
Users of the search or conversational systems often reformulate their inputs when they are dissatisfied with the system responses to their previous inputs. Therefore, reformulation is a useful signal of user dissatisfaction and information related to reformulation has been widely used as a feature to automatically evaluate system performance in web search \cite{Hassan13}, dialogue \cite{Schmitt15}, and IA systems \cite{Jiang15, Kiseleva16a, Sano16}. However, these studies paid little attention to detecting causes of user dissatisfaction. Information of these causes is beneficial to the developers for both improving system design and engineering feature of automatic evaluation methods. Thus, we propose a method to automatically detect the reformulation causes in IAs.

\section{Reformulation Cause Prediction} \label{sec:analysis}
In this section, we describe the task of reformulation cause prediction in IAs.
\begin{table*}[t]
  \begin{center}
    \footnotesize
    \begin{tabular}{llll}
    \textbf{Label} & \textbf{U1} & \textbf{R} & \textbf{U2} \\ \hline
      No error & What's the weather? & It will be sunny today.& What's the weather tomorrow? \\
      ASR error & What's the. & Sorry? & What's the weather? \\
      NLU error & Alarm. & Here are the search results for ``Alarm''.& Open alarm. \\
      LG error & What's your name? & I'm twenty years old. & Tell me your name. \\
      Unsupported action & Play videos of cats. & Sorry, I can't support that action. & Search for videos of cats. \\
      Endpoint error & Search Obama's age.& No results found in for ``Obama's age''. & Search Obama. \\
      Uninterpretable input & Aaaa. & Sorry? & Aaa. \\
    \end{tabular}
    \caption{List of annotation labels. U1, R, and U2 are example conversations. }
    \label{tab:annotation}
  \end{center}
\end{table*}
\begin{table}[t]
  \begin{center}
    \begin{tabular}{lll}
      \textbf{Label} & \textbf{Rate} & \textbf{Error Rate}\\ \hline
      No error & 38.7\% & N.A.\\ \hdashline
      ASR error & 31.7\% & 57.2\% \\
      NLU error & 17.3\% & 31.2\%\\
      LG error & 5.1\% & 9.2\% \\
      Unsupported action & 0.8\% & 1.4\% \\
      Endpoint error & 0.5\% & 0.9\% \\ \hdashline
      Uninterpretable input & 5.9\% & N.A.\\
    \end{tabular}
    \caption{Percentage of annotation labels in our dataset. Error rate is calculated using labels related to reformulation causes.}
    \label{tab:label_percentage}
  \end{center}
\end{table}
\subsection{Definition}
First, we define notations used in this paper.
\begin{description}
  \item[\bf $U_1$]: The user utterance
\item[$R$]: The corresponding system response to $U_1$
\item[$U_2$]: The next utterance to $U_1$
\item[Reformulation]: A pair of ($U_1$, $U_2$) is a \textit{reformulation} if $U_2$ is uttered to modify $U_1$ in order to satisfy the same intent as in $U_1$
\end{description}
To define the \textit{reformulation}, we referred to the definition in the work of \newcite{Hassan15} that is used for voice search systems.

\subsection{Corpus} \label{subsec:corpus}
We constructed a dataset of user logs of a commercial IA\footnote{Because the IA supports only Japanese, all utterances are made in Japanese. In this paper, we present English translations rather than the original Japanese to facilitate non-Japanese readers’ understanding.} for analyzing and predicting reformulation causes. We randomly sampled 1,000 utterance pairs of ($U_1$, $U_2$) and corresponding information with the following conditions.
\begin{itemize}
\item $U_1$ and $U_2$ are text or voice inputs
\item Interval time between $U_1$ and $U_2$ is equal to or less than 30 minutes (the same as in previous research \cite{Jiang15,Sano16}.)
\item Samples where normalized Levenshtein edit-distance \cite{Li07} between $U_1$ and $U_2$ is equal to 0 (i.e., $U_1$ and $U_2$ are identical utterances) or more than 0.5 are excluded
\item $U_1$ and $U_2$ were both uttered in June, 2016
\end{itemize}
With the first three conditions, we can exclude utterances that are not reformulations and can focus on reformulation.
We calculated character-based, rather than word-based, edit-distance (white spaces are ignored), because we found word-based edit-distances sometimes fail to identify reformulation pairs such as "what's up" and "whatsapp."

\begin{figure*}[t]
  \includegraphics[width=0.9\linewidth]{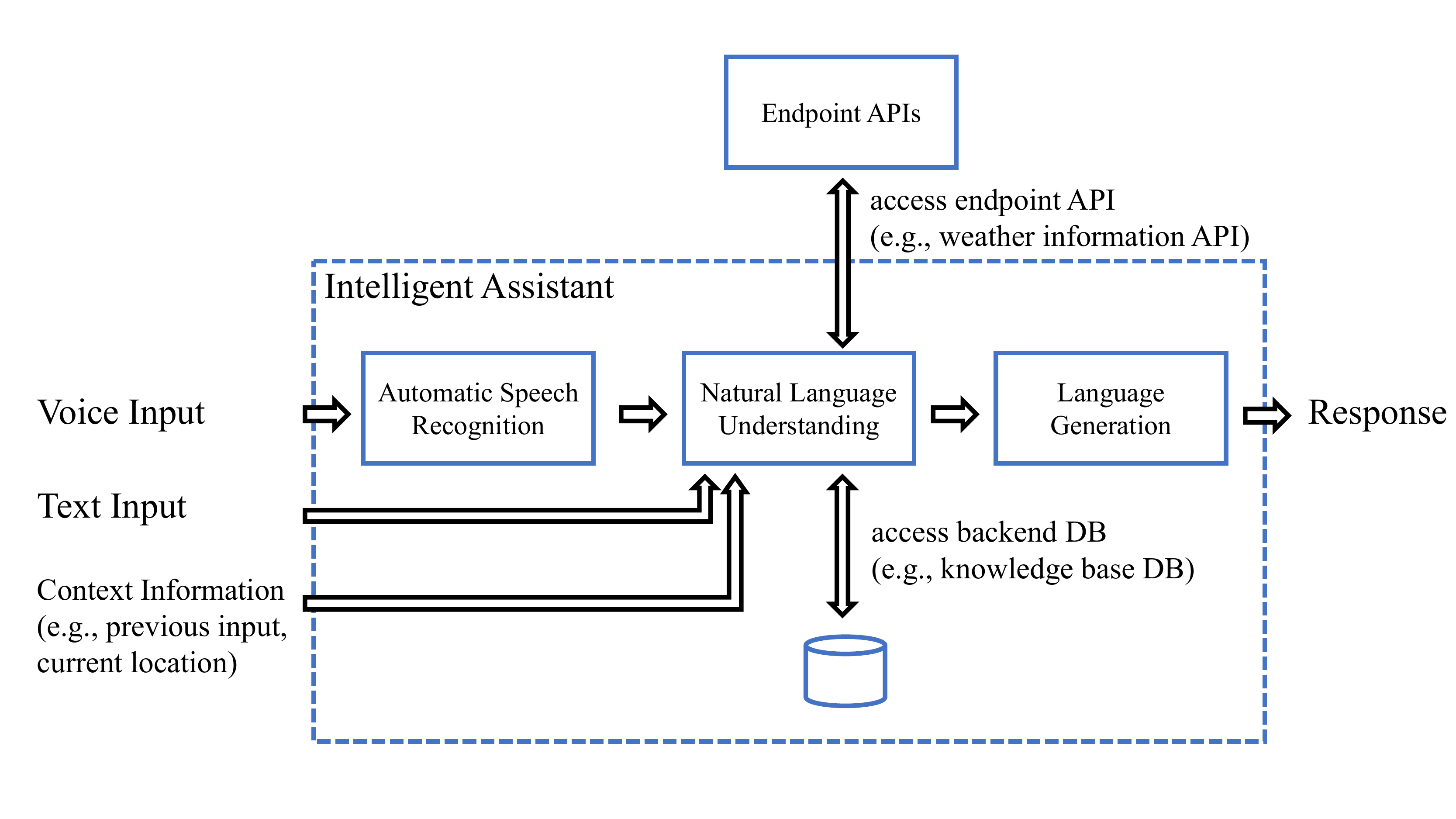}
  \caption{Components and processes of typical IA system.}
 \label{fig:components}
\end{figure*}
All samples in the dataset are manually annotated with the label of the reformulation causes between different components in IAs. Table \ref{tab:annotation} lists the annotation labels. Here, we explain them. In this paper, we assume an IA system that has the components shown in Figure \ref{fig:components}. \textit{ASR error} means that the ASR component misrecognizes $U_1$. \textit{NLU error} means that ASR is correct but the NLU component misunderstands the intent of $U_1$ or fails to fill one or more slots correctly. \textit{LG error} means that ASR and NLU are correct but the LG component fails to generate appropriate response. This error mainly caused by response generation failure in chat intent. Note that the NLU component only determines whether or not an utterance is chat intent and the LG component generates appropriate response of the utterance in our system. Therefore, we consider that the case shown in Table \ref{tab:annotation} is LG error rather than NLU error. \textit{Endpoint error} means that endpoint API (application program interface) fails to respond with correct information. \textit{Unsupported action} means that the system cannot support the action that the user expects and so cannot generate a correct response. \textit{No error} means that a sample contains no error. Submitting similar utterances to obtain better results in search intents (e.g., $U_1$ is ``Search for image of strawberry wallpaper'' and $U_2$ is ``Search for image of strawberry'') and using similar functions (e.g., $U_1$ is ``Turn on Wi-Fi'' and $U_2$ is ``Turn off Wi-Fi'') are typical utterances in this label. Finally, \textit{Uninterpretable input} means that $U_1$ is uninterpretable.

\begin{figure}[t]
  \includegraphics[width=0.9\linewidth]{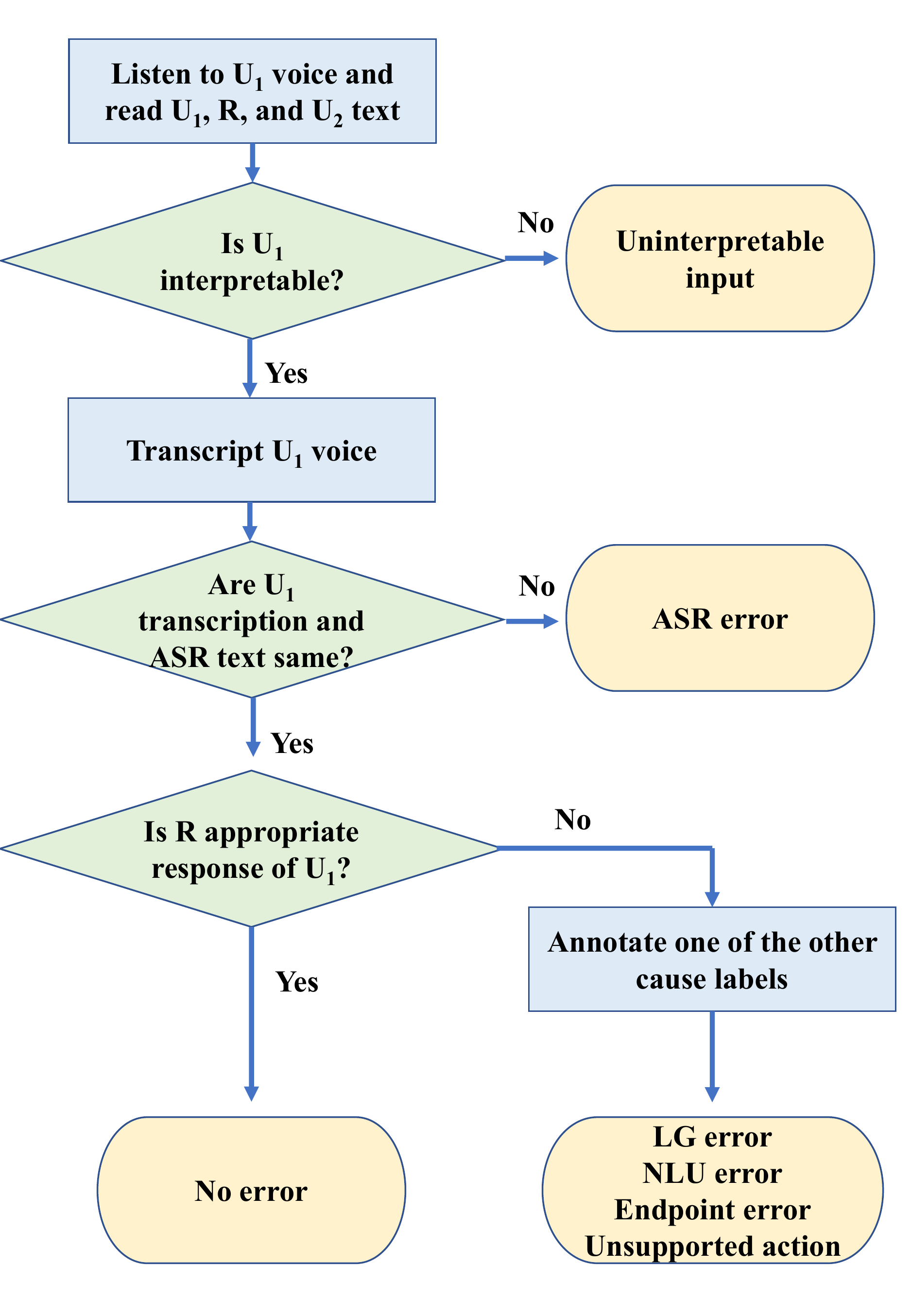}
  \caption{Annotation flowchart.}
 \label{fig:annotation_flow}
\end{figure}

As expert knowledge about the components of IAs is required for the annotation, annotation is performed by an expert developer of the commercial IA.
Figure \ref{fig:annotation_flow} shows the annotation flowchart. First, we listen to the voice of $U_1$ and read the texts of $U_1$, $R$, and $U_2$. Text information is used to support guessing the user intent of $U_1$.
\textit{Uninterpretable input} such as one-word utterances and misrecognized input of background noises is distinguished in this phase. Next, \textit{ASR error} samples are distinguished using transcription of $U_1$.
Afterwards, \textit{No error} samples are distinguished if $R$ of the samples correctly satisfies the intent of $U_1$.
Finally, one of the other four error causes is annotated to the remaining samples.
The annotation results are shown in Table \ref{tab:label_percentage}. We ignored the cases where the latter components could recover from the errors generated by the former components because these cases were rarely observed. For example, a percentage that the NLU component could recover from ASR errors is 1.2\% in our dataset.

\subsection{Discussion on Annotated Labels}
Here, we discuss the annotation results and which labels should be included or excluded in reformulation cause prediction.
To come to the point, we do not use \textit{Uninterpretable input}, \textit{Unsupported action}, or \textit{Endpoint error} for reformulation cause prediction. We will explain why these labels are excluded.

As shown in Table \ref{tab:label_percentage}, the most frequent cause is \textit{ASR error}. This result differs from that of \newcite{Sarikaya17} in which ASR error is the least frequent cause. The reason for the difference is that \newcite{Sarikaya17} used whole samples, whereas we use only reformulation samples. These results indicate that the reformulation tendency when a user encounters an error differs depending on the cause of errors. For example, the percentage of unsupported actions is 1.4\%, which is much smaller than that reported by \newcite{Sarikaya17}, 14\%. This finding indicate that when users encounter a response that notifies them that the action they expected was unsupported, they would rather give up than reformulate their previous utterances. The same is true for endpoint error.

Next, we discuss which labels should be included or excluded in the task. First, \textit{Uninterpretable input} should be excluded because these utterances have no appropriate responses and become noise for the task. We also exclude \textit{Unsupported action} and \textit{Endpoint error}. An IA do not require user feedback for these errors because no components in the IA is responsible for the errors and the IA is aware the error causes in these labels. In addition, the benefits of detecting these errors are limited because these errors are rarely observed in reformulation as shown in Table \ref{tab:label_percentage}. In conclusion, we use \textit{No error}, \textit{ASR error}, \textit{NLU error}, and \textit{LG error} for the task.
\subsection{Task Definition}
Here, we describe the task of reformulation cause prediction.
Our goal is to predict the reformulation cause of $U_1$ between different components using the information from $U_1$ and $U_2$. Specifically, we predict one of the four labels in Table \ref{tab:annotation}, \textit{No error}, \textit{ASR error}, \textit{NLU error}, and \textit{LG error}. For example, we want to predict as \textit{NLU error} from the conversation logs of ($U_1$:``Alarm.'', $R$:``Here are the search results for Alarm.'', $U_2$:``Open alarm.''). These labels are useful for IA developers to improve system performance.

\section{Analysis}
We analyze the statistical differences between reformulation causes prior to the experiments and exploit the findings for engineering features.
\subsection{Analysis of Correction Types} \label{subsec:correction}
\begin{table}[t]
  \begin{center}
    \begin{tabular}{ll}
      \textbf{Type} & \textbf{Example} \\ \hline
      ADD & What's the weather in Nara today?\\
      OMIT & What's the weather? \\
      PAR & How about the weather today? \\
      OTHER & How about the humidity today? \\
    \end{tabular}
    \caption{Example corrections of ``What's the weather today?'' in each correction type.}
    \label{tab:correction_examples}
  \end{center}
\end{table}
\begin{table}[t]
  \begin{center}
    \begin{tabular}{lllll}
       & \textbf{ADD} & \textbf{OMIT} & \textbf{PAR} & \textbf{OTHER} \\ \hline
      No error & 27.1 & 7.2 & 58.1 & 7.4 \\ \hdashline
      ASR error & 7.5 & 6.0 & 74.2 & 12.3 \\
      NLU error & 27.9 & 19.8 & 41.9 & 10.5 \\
      LG error & 23.5 & 19.6 & 47.1 & 9.8 \\
    \end{tabular}
    \caption{Distribution of correction types (\%).}
    \label{tab:correction_types}
  \end{center}
\end{table}
Here, we analyze the correction types when an IA user is trying to correct a previous utterance.
We simplify the definition of \cite{swerts00} and define four correction types: $ADD$, $OMIT$, $PAR$, and $OTHER$.
Their definitions are as follows.
\begin{description}
  \item[ADD] A sequence of words is added to $U_1$.
  \item[OMIT] A sequence of words in $U_1$ is omitted in $U_2$.
  \item[PAR] A sequence of words in $U_1$ changes into another sequence of words in $U_2$.
  \item[OTHER] Match none of preceding correction types.
\end{description}
Note that $U_1$ and $U_2$ have at least one word in common in $ADD$, $OMIT$, and $PAR$.
Table \ref{tab:correction_examples} shows examples of each correction type.

Table \ref{tab:correction_types} presents the distribution of correction types. As shown in Table \ref{tab:correction_types}, the percentage of $PAR$ in \textit{ASR error} is higher than those of other correction types. In \textit{No error}, the percentages of $ADD$ and $PAR$ are large. We can also see that \textit{NLU error} and \textit{LG error} have similar distributions and that  $OMIT$ is more frequent than the other correction types. This is because when users encounter \textit{NLU error} or \textit{LG error}, they use different types of corrections depending on the situation: adding words to clarify their intent, omitting words that seem to be the noise for the system, and so on. We expect correction types to be useful for detecting reformulation causes as the distribution of correction types differs for different error types.

\subsection{Analysis of Input Types} \label{subsec:analysis_input_types}
\label{subsec:input_type}
\begin{table}[t]
  \begin{center}
    \begin{tabular}{lllll}
      & \textbf{V2V} & \textbf{T2T} & \textbf{V2T} & \textbf{T2V} \\ \hline
      No error & 75.2 & 23.5 & 1.3 & 0.0 \\
      ASR error & 94.6 & 0.0 & 5.4 & 0.0 \\
      NLU error & 70.5 & 23.8 & 1.7 & 4.6 \\
      LG error & 76.5 & 23.5 & 0.0 & 0.0 \\
    \end{tabular}
    \caption{Distribution of input type switches (\%). For example, V2T means input type of $U_1$ is voice input and that of $U_2$ is text input.}
    \label{tab:input_types}
  \end{center}
\end{table}
Here, we analyze the correlation between error causes and input types. Input types sometimes differ between $U_1$ and $U_2$. For example, \newcite{Jiang13} have reported that users switch from voice inputs to text inputs when they encounter ASR errors. Table \ref{tab:input_types} presents the distribution of input type switches.
As shown in Table \ref{tab:input_types}, distribution of input type switches differs slightly among the error causes. As \newcite{Jiang13} have shown, voice-to-text switches are more frequent for \textit{ASR error} than in other causes. We can also see that both text-to-voice and voice-to-text switches are observed for \textit{NLU error}. Compared with these causes, input type switches are rarely observed for \textit{No error}. Interestingly, text-to-voice switches are not observed for \textit{No error} either. On the other hand, small number of voice-to-text switches are observed for \textit{No error}. We guess that some users switch input from voice to text when they submit similar query by copy and paste the part of ASR result of their previous utterances. These findings suggest that users tend to keep using the same input type while their intents are correctly recognized. Unlike the other causes, no input type switches are observed for \textit{LG error}. We expect this is due to insufficient data.

\section{Features} \label{sec:features}
\begin{table*}[t]
  \begin{center}
    \begin{tabular}{lll}
      \textbf{Category} & \textbf{Name} & \textbf{Definition}\\ \hline
      Session & \textbf{CharLen*} & Number of characters in utterance. \\
      & \textbf{WordLen*} & Number of words in utterance. \\
      & \textbf{InputType*} & 1 if utterance is text input else 0 \\
      & \textbf{Interval} & Time between $U_1$ and $U_2$ \\ \hdashline
      Reformulation & \textbf{EditDistance} & Normalized Levenshtein edit distance between $U_1$ and $U_2$ \\
      & \textbf{Correction($t$)} & 1 if $U_2$ is correction type $t$ of $U_1$ \\
      & \textbf{CommonWords} & Number of words appearing in both $U_1$ and $U_2$ \\
      & \textbf{Voice2Text} & 1 if $U_1$ is voice input and $U_2$ is text input \\
      & \textbf{Text2Voice} & 1 if $U_1$ is text input and $U_2$ is voice input \\ \hdashline
      ASR & \textbf{ASRConf*} & Speech recognition confidence \\
      & \textbf{VoiceLen*} & Speech recognition time \\ \hdashline
      NLU & \textbf{SameIntent} & 1 if recognized intents between $U_1$ and $U_2$ are same \\
      & \textbf{DifferentIntent} & 1 if recognized intents between $U_1$ and $U_2$ are different \\
      & \textbf{DifferentSlot} & 1 if some slots in $U_2$ are different from those in $U_1$ \\
      & \textbf{IntentType($t$)*} & 1 if recognized intent of utterance is $t$ \\ \hdashline
      LG & \textbf{DialogAct($t$)*} & 1 if utterance contains phrases in dialogue act $t$ \\
    \end{tabular}
    \caption{List of features. Features marked with ``*'' were computed for both $U_1$ and $U_2$.}
    \label{tab:feature}
  \end{center}
\end{table*}
We divide the features into five categories by their functions. Session features and reformulation features, which are useful for reformulation prediction \cite{Hassan15} and system evaluation \cite{Jiang15}, are designed to distinguish between errors or non-errors. Though these features are also useful for reformulation cause detection, these are not specialized for detecting reformulation causes. ASR, NLU, and LG features are designed to detect reformulation causes related to their corresponding components from causes related to the other components. Table \ref{tab:feature} lists the features.
\subsection{Session Features}
Features related to session information belong to this category. In \textit{InputType}, 1 if an utterance is text input and 0 if an utterance is voice input \cite{Hassan15}. In a web search system, the interval time between inputs is a useful indicator of search success \cite{Huang09}. Therefore, \textit{Interval} is useful for distinguishing \textit{No error} from the other labels. If \textit{CharLen} or \textit{WordLen} of the utterance is long or short, the utterance possibly contains noise information or lacks information for the system.
\subsection{Reformulation Features}
Features related to reformulation belong to this category. Features in this category are widely used in previous methods such as query reformulation detection \cite{Hassan15}, error detection \cite{Litman06}, and system performance evaluation \cite{Jiang15}. The correction type $t$ of \textit{Correction($t$)} is one of ADD, OMIT, PAR, or OTHER described in Section \ref{subsec:correction}. As shown in Section \ref{subsec:correction}, the distribution of correction types differs among annotation labels. Therefore, \textit{Correction($t$)} is useful information for predicting reformulation causes. \textit{Voice2Text} and \textit{Text2Voice} are designed to distinguish \textit{ASR error} and \textit{NLU error} from other errors on the basis of analysis in Section \ref{subsec:input_type}.
\subsection{ASR Features}
Features related to the ASR component belong to this category. Low \textit{ASRConf} indicates speech recognition errors \cite{Hassan15}. The probability of misrecognition increases as the recognized voice length increases \cite{Hassan15}. Therefore, long \textit{VoiceLen} may be one signal related to ASR error. Note that these features are calculated only when the input type of the utterance is voice input.
\subsection{NLU Features}
Features related to the NLU component belong to this category. When a user's intent in $U_1$ is misunderstood and that in $U_2$ is correctly understood, recognized intents or filled slots between the utterances are different. On the other hand, when a user's intent in $U_1$ is correctly understood by the system, the user sometimes uses similar functions subsequently (e.g., requesting weather information for Osaka after requesting it for Tokyo.). Therefore, \textit{DifferentIntent} and \textit{DifferentSlot} are useful to distinguish \textit{NLU error} from the other errors, and \textit{SameIntent} is useful to distinguish \textit{No error} from the other errors. Note that the intents used in these features are the intents recognized by the IA system such as weather information, web search, application launch, and chat.
\subsection{LG Features}
\begin{table}[t]
  \begin{center}
    \begin{tabular}{ll}
      \textbf{Type} & \textbf{Examples} \\ \hline
      Praise & Wow!; Great. \\
      Thanking & Thanks.; Thank you. \\
      Backchannel & I see.; Yeah. \\
      Accept & Yes.; Exactly. \\
      Abuse & Shit.; Shut up. \\
      Reject & No.; Not like that. \\
      IDU & What do you mean? \\
    \end{tabular}
    \caption{List of dialog acts.}
    \label{tab:dialog_act}
  \end{center}
\end{table}
Features related to the LG component belong to this category. If users expect the system to chat, their utterances may contain phrases commonly used in chatting. \textit{DialogAct($t$)} is designed to detect these phrases in the utterance. User utterances of chat intent in IAs have unique characterictics (e.g., some users curse at the intelligent assistants \cite{Akasaki17}). We defined seven types of dialogue acts that are common in chats between users and IAs, as listed in Table \ref{tab:dialog_act}. Frequently occurring phrases in the user log of the commercial IA are used for the phrases of each dialogue act.

\section{Experiments}
\subsection{Experimental Settings}
The experimental settings of reformulation cause prediction are as follows.
\begin{itemize}
  \item The dataset described in Section \ref{subsec:corpus} is used for evaluation. It contains 928 samples.
  \item We evaluate the performance of the model with 10-fold cross validation.
  \item We train the model using a linear SVM classifier with the features described in section \ref{sec:features}.
  \item We optimize hyper parameters of the classifier with an additional 5-fold cross validation using only training sets (9-folds used for a training set is combined and split into 5 new folds in each validation).
  \item The baseline model is trained in the same conditions except that the model uses only Session and Reformulation features. Comparison between the proposed and the baseline methods enables us to evaluate the effect of the features related to the components of IA.
\end{itemize}
We choose linear SVM because it has scalability and has outperformed RBF-kernel SVM. 
\subsection{Results} \label{subsec:results}
\begin{table}[t]
  \begin{center}
    \begin{tabular}{llll}
      &  \textbf{Precision} & \textbf{Recall} & \textbf{F$_1$} \\ \hline
      Baseline (B.) & 0.51 & 0.53 & 0.51 \\
      Proposed & \textbf{0.67} & \textbf{0.66} & \textbf{0.67}$^+$ \\ \hdashline
      B. + ASR & 0.56 & 0.59 & 0.57$^+$ \\
      B. + NLU & 0.63 & 0.60 & 0.61$^\star$ \\
      B. + LG & 0.50 & 0.50 & 0.49 \\
    \end{tabular}
    \caption{The results of the reformulation cause prediction. }
    \label{tab:exp_overall}
  \end{center}
\end{table}
Table \ref{tab:exp_overall} presents the results for the reformulation cause prediction. The first row compares the proposed method with the baseline. The proposed method obtains a 0.67-point F$_1$-measure and outperforms the baseline. This result shows the effectiveness of the features related to the components of IA. The second row illustrates the performance when one feature set is added to the baseline. We can see that ASR and NLU features improve the performance of the baseline. In F$_1$-measure, statistical significant differences from the baseline detected by the paired t-test are denoted by $+$ ($p < 0.01$) and $\star$ ($p < 0.05$).
\begin{table}[t]
  \begin{center}
    \begin{tabular}{lcccc}
      Gold \textbackslash \ Predict &  No & ASR & NLU & LG \\ \hline
      No error & \textbf{216} & 132 & 29 & 10\\
      ASR error & 67 & \textbf{219} & 21 & 10 \\
      NLU error & 61 & 54 & \textbf{52} & 6 \\
      LG error & 19 & 17 & 14 & \textbf{1} \\
    \end{tabular} \\
    (a) Baseline \\
    \begin{tabular}{lcccc}
      Gold \textbackslash \ Predict &  No & ASR & NLU & LG \\ \hline
      No error & \textbf{284} & 55 & 27 & 21 \\
      ASR error & 38 & \textbf{230} & 37 & 12 \\
      NLU error & 44 & 29 & \textbf{81} & 19 \\
      LG error & 8 & 12 & 11 & \textbf{20} \\
    \end{tabular} \\
    (b) Proposed \\
    \caption{Confusion matrix of reformulation cause prediction of (a) baseline and (b) proposed methods.}
    \label{tab:exp_overall_cm}
  \end{center}
\end{table}

Table \ref{tab:exp_overall_cm} presents the confusion matrix of the proposed and baseline methods. The results going diagonally show agreement between the gold labels and the predicted labels. As shown in Table \ref{tab:exp_overall_cm}, the proposed method outperforms the baseline regardless of the gold labels. Again, these results indicate the effectiveness of the features related to the components of IA.

Table \ref{tab:exp_components} presents F$_1$-measures of each gold label. Again, the proposed method outperforms the other methods. Focusing on individual labels, F$_1$-measures of the proposed method are better for \textit{No error} and \textit{ASR error} than for \textit{NLU error} and \textit{LG error}. In other words, F$_1$-measures for \textit{NLU error} and \textit{LG error} are not high. As the performances of individual labels are in the order of the number of samples belonging to the label, we expect that these low performances are mainly due to insufficient data and will improve given sufficient data. Focusing on individual feature sets, ASR and NLU features are useful for distinguishing \textit{No error} from other labels. Note that statistical significant differences from the baseline detected by the paired t-test are denoted by $+$ ($p < 0.01$) and $\star$ ($p < 0.05$).
\begin{table}[t]
  \begin{center}
    \begin{tabular}{lllll}
      & \textbf{No} & \textbf{ASR} & \textbf{NLU} & \textbf{LG} \\ \hline
      Baseline (B.) & 0.58 & 0.59 & 0.36 & 0.03 \\
      Proposed & \textbf{0.75}$^+$ & \textbf{0.72}$^+$ & \textbf{0.49}$^\star$ & \textbf{0.33}$^+$ \\ \hdashline
      B. + ASR & 0.66$^+$ & 0.67$^+$ & 0.35 & 0.16 \\
      B. + NLU & 0.71$^+$ & 0.65 & 0.43 & 0.25$^\star$ \\
      B. + LG & 0.55 & 0.57 & 0.32 & 0.08 \\
    \end{tabular}
    \caption{F$_1$-measure in reformulation cause prediction for each label.}
    \label{tab:exp_components}
  \end{center}
\end{table}
\subsection{Results by Input Types}
Here we analyze the result of Table \ref{tab:exp_overall} by input types.
Table \ref{tab:exp_input_types} presents the results of reformulation cause prediction by input types.
As shown in Table \ref{tab:exp_input_types}, Both the F$_1$-measures of the proposed method in voice inputs and text inputs are 0.66. These results suggest that the proposed method is robust for both input types. On the other hand, the F$_1$-measure of the baseline method in voice inputs is lower than that in text inputs. Particularly, the F$_1$ measure of \textit{No error} in voice inputs is lower than that in text inputs. Table \ref{tab:exp_input_types_filtered} presents the distribution of predicted labels with the following two conditions. First, $U_1$ and $U_2$ are voice inputs. Second, gold labels of all samples are \textit{No error}. As shown in Table \ref{tab:exp_input_types_filtered}, misclassification rate of the proposed method in \textit{No error} as \textit{ASR error} is less than that of the baseline method. In other words, the proposed method distinguishes between \textit{No error} and \textit{ASR error} more accurately compared to the baseline method. These results suggest that ASR features contribute to the performance improvement of the proposed method.
\begin{table}[t]
  \begin{center}
    \begin{tabular}{lccccc}
      & No & ASR & NLU & LG & total \\ \hline
      Baseline & 0.49 & 0.58 & 0.33 & 0.03 & 0.47 \\
      Proposed & \textbf{0.73} & \textbf{0.71} & \textbf{0.49} & \textbf{0.30} & \textbf{0.66} \\ \hdashline
      \# samples & 291 & 300 & 122 & 39 & 752 \\
    \end{tabular} \\
    (a) $U_1$ and $U_2$ are voice inputs. \\
    \begin{tabular}{lccccc}
      & No & ASR & NLU & LG & total \\ \hline
      Baseline & 0.80 & N.A. & 0.32 & 0.00 & 0.60 \\
      Proposed & \textbf{0.81} & N.A. & \textbf{0.42} & \textbf{0.39} & \textbf{0.66} \\ \hdashline
      \# samples & 91 & N.A. & 40 & 12 & 143 \\
    \end{tabular} \\
    (b) $U_1$ and $U_2$ are text inputs. \\
    \caption{F$_1$-measure in reformulation cause prediction of each label between input types. }
    \label{tab:exp_input_types}
  \end{center}
\end{table}

\begin{table}[t]
  \begin{center}
    \begin{tabular}{lcccc}
      Predicted label & No & ASR & NLU & LG \\ \hline
      Baseline & 129 & 129 & 25 & 8 \\
      Proposed & 205 & 53 & 16 & 17 \\
    \end{tabular}
    \caption{ The number of predicted samples in reformulation cause prediction in following two conditions. First, $U_1$ and $U_2$ are voice inputs. Second, gold labels of all samples are \textit{No error}. }
    \label{tab:exp_input_types_filtered}
  \end{center}
\end{table}

\subsection{Investigation of Feature Weights}
\begin{table}[t]
  \footnotesize
  \begin{center}
    \begin{tabular}{llc}
      Label & Feature & Weight \\ \hline
      No error & ASRConf* & 1.07 \\
               & IntentType(SingSong)* & 1.05 \\ \hdashline
               & Voice2Text & -1.01 \\
               & IntentType(DeviceControl)* & -1.10 \\ \hline
      ASR error & Voice2Text & 1.27 \\
                & IntentType(Search)* & 0.90 \\ \hdashline
                & ASRConf* & -1.51 \\
                & InputType* & -1.77 \\ \hline
      NLU error & Text2Voice & 1.43 \\
                & InputType* & 0.94 \\ \hdashline
                & SameIntent & -0.61 \\
                & IntentType(Dictionary)* & -0.77 \\ \hline
      LG error & IntentType(DeviceControl)* & 1.08 \\
               & DialogAct(IDU)* & 0.81 \\ \hdashline
               & DialogAct(Praise)* & -0.95 \\
               & Voice2Text & -1.04 \\
    \end{tabular}
  \caption{ Top and Bottom two feature weights of the proposed method. Features marked with ``*'' were computed for $U_1$. }
    \label{tab:feature_weights}
  \end{center}
\end{table}
We investigate weights of the features learned by the linear-kernel SVM to clarify what features contribute to the reformulation cause prediction.

Table \ref{tab:feature_weights} presents top and bottom feature weights in each label. The median value of the 10 models which are obtained with cross validation are used for weights in Table \ref{tab:feature_weights}.
Features calculated from $U_1$ appear in Table \ref{tab:feature_weights} but that calculated from $U_2$ do not appear. This result is not surprising because information related to $U_1$ has more relationship to reformulation causes compared to that related to $U_2$.
Next, we focus on the features in individual labels. Features related to their corresponding components appear in Table \ref{tab:feature_weights} such as \textit{ASRConf} in \textit{ASR error}, \textit{SameIntent} in \textit{NLU error}, and \textit{DialogAct} in \textit{LG error}. These results indicate that features designed to detect reformulation causes related to their corresponding components work as designed.
Finally, we focus on the individual features. We observe that features of input type switches are useful for predicting reformulation causes. In particular, \textit{Voice2Text} is useful for detecting \textit{ASR error} and \textit{Text2Voice} is useful for detecting \textit{NLU error}. These results are consistent with findings in section \ref{subsec:analysis_input_types}.

%

\section{Future Work}
While the proposed method has outperformed the baseline, there is room for improvement on the performance.
As mentioned in section \ref{subsec:results}, the performances of individual labels are in the order of the number of samples belonging to the labels. Therefore, we expect that the performance improves as the dataset size increases.
The performance will also improve if some features used in previous studies are added to our features. For example, linguistic features such as word n-gram and language model score are used in previous studies \cite{Hassan15, Meena15} but not used in ours.
%

\section{Conclusion}
This paper attempted to predict reformulation causes in intelligent assistants (IAs). Prior to the prediction, we first analyzed the cause of reformulation in IAs using user logs obtained from a commercial IA. Based on the analysis, we defined reformulation cause prediction as four-class classification problem of classifying user utterances into \textit{No error}, \textit{ASR error}, \textit{NLU error}, or \textit{LG error}. Features are divided into five categories mainly on the basis of the relations with the components in the IA.
The experiments demonstrated that the proposed method, which combines all feature sets, outperforms the baseline which uses component-independent features such as session information and reformulation related information.

\bibliography{sigdial2017}

\begin{thebibliography}{}
\expandafter\ifx\csname natexlab\endcsname\relax\def\natexlab#1{#1}\fi

\bibitem[{Akasaki and Kaji(2017)}]{Akasaki17}
Satoshi Akasaki and Nobuhiro Kaji. 2017.
\newblock Chat detection in an intelligent assistant: Combining task-oriented
  and non-task-oriented spoken dialogue systems.
\newblock In {\em Proceedings of the 55th Annual Meeting of the Association for
  Computational Linguistics (Volume 1: Long Papers)\/}. Association for
  Computational Linguistics (to appear).

\bibitem[{Feild et~al.(2010)Feild, Allan, and Jones}]{Feild10}
Henry~A. Feild, James Allan, and Rosie Jones. 2010.
\newblock \href{https://doi.org/10.1145/1835449.1835458}{Predicting searcher
  frustration}.
\newblock In {\em Proceedings of the 33rd International ACM SIGIR Conference on
  Research and Development in Information Retrieval\/}. ACM, SIGIR '10, pages
  34--41.
\newblock
  \href{https://doi.org/10.1145/1835449.1835458}{https://doi.org/10.1145/1835449.1835458}.

\bibitem[{Georgiladakis et~al.(2016)Georgiladakis, Athanasopoulou, Meena,
  Lopes, Chorianopoulou, Palogiannidi, Iosif, Skantze, and
  Potamianos}]{Georgiladakis16}
Spiros Georgiladakis, Georgia Athanasopoulou, Raveesh Meena, Jos^^c3^^a9 Lopes,
  Arodami Chorianopoulou, Elisavet Palogiannidi, Elias Iosif, Gabriel Skantze,
  and Alexandros Potamianos. 2016.
\newblock \href{https://doi.org/10.21437/Interspeech.2016-1273}{Root cause
  analysis of miscommunication hotspots in spoken dialogue systems}.
\newblock In {\em Proceedings of Interspeech 2016\/}. ISCA, pages 1156--1160.
\newblock
  \href{https://doi.org/10.21437/Interspeech.2016-1273}{https://doi.org/10.21437/Interspeech.2016-1273}.

\bibitem[{Hassan et~al.(2015)Hassan, Gurunath~Kulkarni, Ozertem, and
  Jones}]{Hassan15}
Ahmed Hassan, Ranjitha Gurunath~Kulkarni, Umut Ozertem, and Rosie Jones. 2015.
\newblock \href{https://doi.org/10.1145/2806416.2806491}{Characterizing and
  predicting voice query reformulation}.
\newblock In {\em Proceedings of the 24th ACM International on Conference on
  Information and Knowledge Management\/}. ACM, pages 543--552.
\newblock
  \href{https://doi.org/10.1145/2806416.2806491}{https://doi.org/10.1145/2806416.2806491}.

\bibitem[{Hassan et~al.(2013)Hassan, Shi, Craswell, and Ramsey}]{Hassan13}
Ahmed Hassan, Xiaolin Shi, Nick Craswell, and Bill Ramsey. 2013.
\newblock \href{https://doi.org/10.1145/2505515.2505682}{Beyond clicks: Query
  reformulation as a predictor of search satisfaction}.
\newblock In {\em Proceedings of the 22nd ACM International Conference on
  Information and Knowledge Management\/}. ACM, pages 2019--2028.
\newblock
  \href{https://doi.org/10.1145/2505515.2505682}{https://doi.org/10.1145/2505515.2505682}.

\bibitem[{Hirschberg et~al.(2001)Hirschberg, Litman, and
  Swerts}]{Hirschberg01a}
Julia Hirschberg, Diane Litman, and Marc Swerts. 2001.
\newblock \href{https://doi.org/10.3115/1073336.1073363}{Identifying user
  corrections automatically in spoken dialogue systems}.
\newblock In {\em Proceedings of the Second Meeting of the North American
  Chapter of the Association for Computational Linguistics on Language
  Technologies\/}. Association for Computational Linguistics, NAACL '01, pages
  1--8.
\newblock
  \href{https://doi.org/10.3115/1073336.1073363}{https://doi.org/10.3115/1073336.1073363}.

\bibitem[{Hirst et~al.(1994)Hirst, McRoy, Heeman, Edmonds, and
  Horton}]{Hirst94}
Graeme Hirst, Susan McRoy, Peter Heeman, Philip Edmonds, and Diane Horton.
  1994.
\newblock \href{https://doi.org/10.1016/0167-6393(94)90073-6}{Repairing
  conversational misunderstandings and non-understandings}.
\newblock {\em Speech Communication\/} 15(3-4):213--229.
\newblock
  \href{https://doi.org/10.1016/0167-6393(94)90073-6}{https://doi.org/10.1016/0167-6393(94)90073-6}.

\bibitem[{Huang and Efthimiadis(2009)}]{Huang09}
Jeff Huang and Efthimis~N. Efthimiadis. 2009.
\newblock \href{https://doi.org/10.1145/1645953.1645966}{Analyzing and
  evaluating query reformulation strategies in web search logs}.
\newblock In {\em Proceedings of the 18th ACM Conference on Information and
  Knowledge Management\/}. ACM, pages 77--86.
\newblock
  \href{https://doi.org/10.1145/1645953.1645966}{https://doi.org/10.1145/1645953.1645966}.

\bibitem[{Jiang et~al.(2015)Jiang, Hassan~Awadallah, Jones, Ozertem, Zitouni,
  Gurunath~Kulkarni, and Khan}]{Jiang15}
Jiepu Jiang, Ahmed Hassan~Awadallah, Rosie Jones, Umut Ozertem, Imed Zitouni,
  Ranjitha Gurunath~Kulkarni, and Omar~Zia Khan. 2015.
\newblock \href{https://doi.org/10.1145/2736277.2741669}{Automatic online
  evaluation of intelligent assistants}.
\newblock In {\em Proceedings of the 24th International Conference on World
  Wide Web\/}. International World Wide Web Conferences Steering Committee,
  pages 506--516.
\newblock
  \href{https://doi.org/10.1145/2736277.2741669}{https://doi.org/10.1145/2736277.2741669}.

\bibitem[{Jiang et~al.(2013)Jiang, Jeng, and He}]{Jiang13}
Jiepu Jiang, Wei Jeng, and Daqing He. 2013.
\newblock \href{https://doi.org/10.1145/2484028.2484092}{How do users respond
  to voice input errors?: Lexical and phonetic query reformulation in voice
  search}.
\newblock In {\em Proceedings of the 36th International ACM SIGIR Conference on
  Research and Development in Information Retrieval\/}. ACM, pages 143--152.
\newblock
  \href{https://doi.org/10.1145/2484028.2484092}{https://doi.org/10.1145/2484028.2484092}.

\bibitem[{Kiseleva et~al.(2016)Kiseleva, Williams, Jiang, Hassan~Awadallah,
  Crook, Zitouni, and Anastasakos}]{Kiseleva16a}
Julia Kiseleva, Kyle Williams, Jiepu Jiang, Ahmed Hassan~Awadallah, Aidan~C.
  Crook, Imed Zitouni, and Tasos Anastasakos. 2016.
\newblock \href{https://doi.org/10.1145/2854946.2854961}{Understanding user
  satisfaction with intelligent assistants}.
\newblock In {\em Proceedings of the 2016 ACM on Conference on Human
  Information Interaction and Retrieval\/}. ACM, pages 121--130.
\newblock
  \href{https://doi.org/10.1145/2854946.2854961}{https://doi.org/10.1145/2854946.2854961}.

\bibitem[{Levow(1998)}]{Levow98}
Gina-Anne Levow. 1998.
\newblock \href{https://doi.org/10.3115/980845.980969}{Characterizing and
  recognizing spoken corrections in human-computer dialogue}.
\newblock In {\em Proceedings of the 36th Annual Meeting of the Association for
  Computational Linguistics and 17th International Conference on Computational
  Linguistics - Volume 1\/}. Association for Computational Linguistics, ACL
  '98, pages 736--742.
\newblock
  \href{https://doi.org/10.3115/980845.980969}{https://doi.org/10.3115/980845.980969}.

\bibitem[{Li and Liu(2007)}]{Li07}
Yujian Li and Bo~Liu. 2007.
\newblock \href{https://doi.org/10.1109/TPAMI.2007.1078}{A normalized
  {L}evenshtein distance metric}.
\newblock {\em Pattern Analysis and Machine Intelligence, IEEE Transactions
  on\/} 29(6):1091--1095.
\newblock
  \href{https://doi.org/10.1109/TPAMI.2007.1078}{https://doi.org/10.1109/TPAMI.2007.1078}.

\bibitem[{Litman et~al.(2006)Litman, Hirschberg, and Swerts}]{Litman06}
Diane Litman, Julia Hirschberg, and Marc Swerts. 2006.
\newblock \href{https://doi.org/10.1162/coli.2006.32.3.417}{Characterizing and
  predicting corrections in spoken dialogue systems}.
\newblock {\em Computational Linguistics\/} 32(3):417--438.
\newblock
  \href{https://doi.org/10.1162/coli.2006.32.3.417}{https://doi.org/10.1162/coli.2006.32.3.417}.

\bibitem[{Meena et~al.(2015)Meena, Lopes, Skantze, and Gustafson}]{Meena15}
Raveesh Meena, Jose Lopes, Gabriel Skantze, and Joakim Gustafson. 2015.
\newblock \href{https://doi.org/10.18653/v1/W15-4647}{Automatic detection of
  miscommunication in spoken dialogue systems}.
\newblock In {\em Proceedings of the 16th Annual Meeting of the Special
  Interest Group on Discourse and Dialogue\/}. Association for Computational
  Linguistics, pages 354--363.
\newblock
  \href{https://doi.org/10.18653/v1/W15-4647}{https://doi.org/10.18653/v1/W15-4647}.

\bibitem[{Raux et~al.(2005)Raux, Langner, Bohus, Black, and Eskenazi}]{Raux05}
Antoine Raux, Brian Langner, Dan Bohus, Alan~W Black, and Maxine Eskenazi.
  2005.
\newblock Let’s go public! taking a spoken dialog system to the real world.
\newblock In {\em Proceedings of Interspeech 2005\/}. ISCA, pages 885--888.

\bibitem[{Sano et~al.(2016)Sano, Kaji, and Sassano}]{Sano16}
Shumpei Sano, Nobuhiro Kaji, and Manabu Sassano. 2016.
\newblock \href{https://doi.org/10.18653/v1/P16-1114}{Prediction of prospective
  user engagement with intelligent assistants}.
\newblock In {\em Proceedings of the 54th Annual Meeting of the Association for
  Computational Linguistics (Volume 1: Long Papers)\/}. Association for
  Computational Linguistics, pages 1203--1212.
\newblock
  \href{https://doi.org/10.18653/v1/P16-1114}{https://doi.org/10.18653/v1/P16-1114}.

\bibitem[{Sarikaya(2017)}]{Sarikaya17}
Ruhi Sarikaya. 2017.
\newblock \href{https://doi.org/10.1109/MSP.2016.2617341}{The technology behind
  personal digital assistants: An overview of the system architecture and key
  components}.
\newblock {\em IEEE Signal Processing Magazine\/} 34(1):67--81.
\newblock
  \href{https://doi.org/10.1109/MSP.2016.2617341}{https://doi.org/10.1109/MSP.2016.2617341}.

\bibitem[{Schmitt and Ultes(2015)}]{Schmitt15}
Alexander Schmitt and Stefan Ultes. 2015.
\newblock \href{https://doi.org/10.1016/j.specom.2015.06.003}{Interaction
  quality: Assessing the quality of ongoing spoken dialog interaction by
  experts―and how it relates to user satisfaction}.
\newblock {\em Speech Commun.\/} 74(C):12--36.
\newblock
  \href{https://doi.org/10.1016/j.specom.2015.06.003}{https://doi.org/10.1016/j.specom.2015.06.003}.

\bibitem[{Shokouhi et~al.(2016)Shokouhi, Ozertem, and Craswell}]{Shokouhi16}
Milad Shokouhi, Umut Ozertem, and Nick Craswell. 2016.
\newblock \href{https://doi.org/10.1145/2872427.2882994}{Did you say u2 or
  youtube?: Inferring implicit transcripts from voice search logs}.
\newblock In {\em Proceedings of the 25th International Conference on World
  Wide Web\/}. International World Wide Web Conferences Steering Committee, WWW
  '16, pages 1215--1224.
\newblock
  \href{https://doi.org/10.1145/2872427.2882994}{https://doi.org/10.1145/2872427.2882994}.

\bibitem[{Swerts et~al.(2000)Swerts, Litman, and Hirschberg}]{swerts00}
Marc Swerts, Diane~J Litman, and Julia Hirschberg. 2000.
\newblock Corrections in spoken dialogue systems.
\newblock In {\em Proceedings of ICSLP-2000\/}. ISCA, pages 615--618.

\end{thebibliography}
\bibliographystyle{acl_natbib}

\appendix

\end{document}